\documentclass{article}

\usepackage[nonatbib,preprint]{neurips_2024}

\newenvironment{tightlist}%
{\begin{list}{$\bullet$}{%
    \setlength{\topsep}{0in}
    \setlength{\partopsep}{0in}
    \setlength{\itemsep}{0in}
    \setlength{\parsep}{0in}
    \setlength{\leftmargin}{1.5em}
    \setlength{\rightmargin}{0in}
}
}%
{\end{list}
}

\usepackage[utf8]{inputenc} 
\usepackage[T1]{fontenc}    
\usepackage{hyperref}       
\usepackage{url}            
\usepackage{booktabs}       
\usepackage{amsfonts}       
\usepackage{nicefrac}       
\usepackage{microtype}      
\usepackage{xcolor}         
\usepackage{algorithm}
\usepackage{algpseudocode}
\usepackage{subfigure}
\usepackage{xcolor}
\usepackage{verbatim}
\usepackage{fancyvrb}
\usepackage{graphicx} 
\usepackage{amsmath}
\usepackage{wrapfig}

\usepackage{graphics} 
\usepackage{epsfig} 
\usepackage{times} 
\usepackage{amsmath} 
\usepackage{amssymb}  
\usepackage{booktabs}

\title{Learning to Bridge the Gap: Efficient Novelty Recovery with Planning and Reinforcement Learning}

\author{Alicia Li~\thanks{Equal Contribution.}, Nishanth Kumar$^{*}$, Tom\'as Lozano-P\'erez, and Leslie Pack Kaelbling\\
MIT CSAIL\\
\{\texttt{aliciali, njk, tlp, lpk}\}@mit.edu
}

\begin{document}

\maketitle
\thispagestyle{empty}
\pagestyle{empty}

\begin{abstract}
The real world is unpredictable. Therefore, to solve long-horizon decision-making problems with autonomous robots, we must construct agents that are capable of adapting to changes in the environment during deployment. Model-based planning approaches can enable robots to solve complex, long-horizon tasks in a variety of environments. However, such approaches tend to be brittle when deployed into an environment featuring a novel situation that their underlying model does not account for. In this work, we propose to learn a ``bridge policy'' via Reinforcement Learning (RL) to adapt to such novelties. We introduce a simple formulation for such learning, where the RL problem is constructed with a special ``CallPlanner'' action that terminates the bridge policy and hands control of the agent back to the planner. This allows the RL policy to learn the set of states in which querying the planner and following the returned plan will achieve the goal. We show that this formulation enables the agent to rapidly learn by leveraging the planner's knowledge to avoid challenging long-horizon exploration caused by sparse reward. In experiments across three different simulated domains of varying complexity, we demonstrate that our approach is able to learn policies that adapt to novelty more efficiently than several baselines, including a pure RL baseline. We also demonstrate that the learned bridge policy is generalizable in that it can be combined with the planner to enable the agent to solve more complex tasks with multiple instances of the encountered novelty.
\end{abstract}


\section{Introduction}
\label{sec:intro}
Recent model-based planning approaches, such as Task and Motion Planning (TAMP), have enabled robots to perform complex and long-horizon tasks, such as setting a table, rearranging a room, or even assembling complex structures, under a wide variety of circumstances~\cite{tamp_survey,kumar2024practice,chen2022cooperativetaskmotionplanning}.
These approaches assume access to some form of structured, abstract \textit{model} that captures aspects of the environment important for the robot's decision making.
However, useful robots must be able to cope with \textit{novelty} that arises from being deployed in  environments that neither the robot nor its designers could possibly have seen ahead of time~\cite{kumar2024practice}. 
Unfortunately, following the planner's proposed action sequence fails catastrophically in such cases, which significantly limits the applicability and utility of such model-based approaches.
We are thus primarily interested in efficiently and autonomously learning to cope with environment novelties encountered during an agent's deployment in a novel environment.

\begin{figure*}[!ht]
\centering
\subfigure[Encountering a Failure]{
\scalebox{0.25}{\includegraphics[]{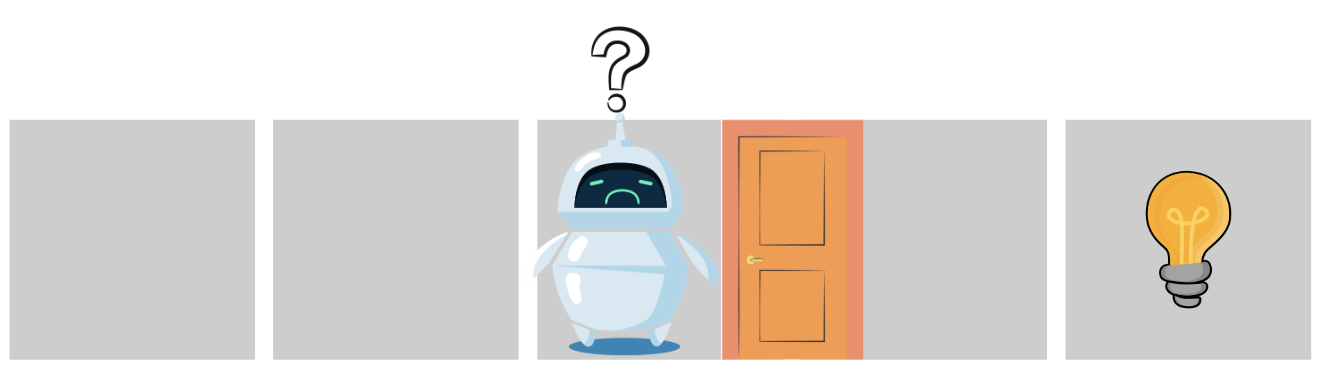}}
}
\centering
\subfigure[Learning to Recover]{
\scalebox{0.25}{\includegraphics[]{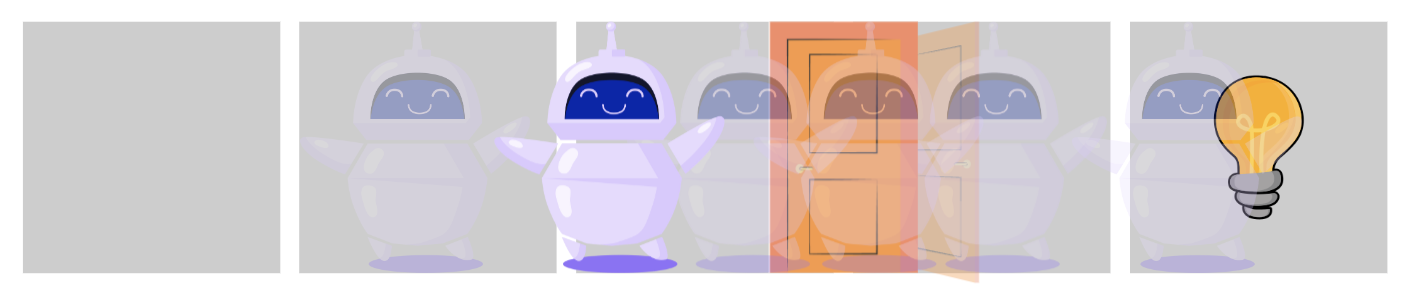}}
}
\centering
\subfigure[Deploying RL Bridge Policy Approach]{
\scalebox{0.4}{\includegraphics[]{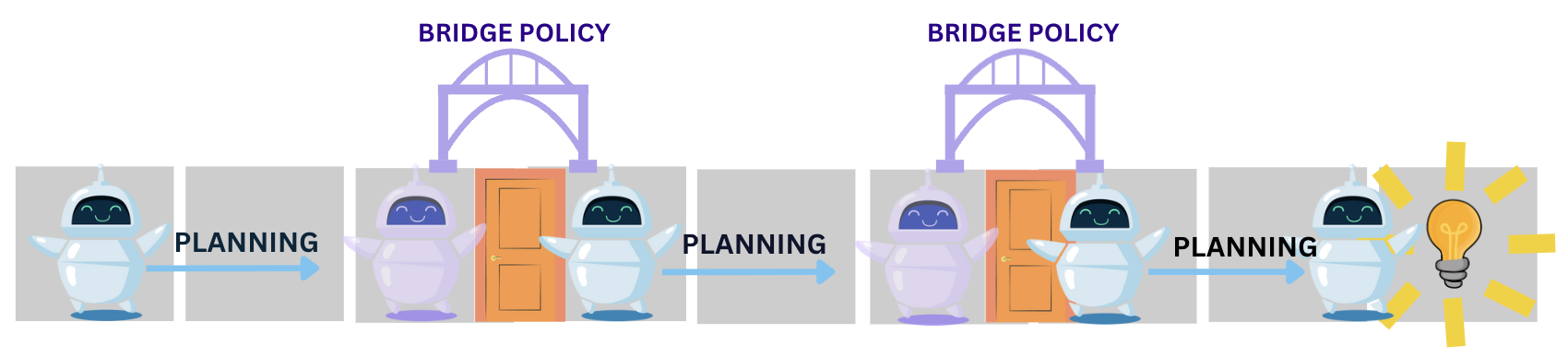}}
}
\caption{\small{Our approach in Light Switch Door. (a): During deployment, the robot encounters an unknown object (door), when trying to turn on the light. This renders it unable to follow its plan.
 (b): The robot switches from planning to RL to learn how to overcome the novelty.
 (c): During evaluation, the robot switches from following its plan to following the learned `bridge policy' each time it gets stuck at a door, and then switches back to the planner once it opens the door. It is thus able to generalize to opening an arbitrary number of doors.}}
\label{fig:overview}
\vspace{-1em}
\end{figure*}

In this work, we assume the robot is equipped with a collection of \textit{skills} (such as moving to a specified location, picking a specified object, or placing a held object at a location) as well as a planning model in terms of those skills that it can use to solve user-provided tasks.
We are interested in situations where executing plans made according to the model leads to some kind of failure in the environment.
As a simple example, consider the `Light Switch Door' environment illustrated in Figure~\ref{fig:overview}.
Here, the robot possesses skills to move left and right between grid cells, and turn on and off the light once it is in the same cell.
When provided with the task of turning on the light, it is able to make and execute a plan composed of these skills to move right until it is in the light's cell, and then turn it on.
However, the robot is deployed into an environment containing one or more doors that prevent it from moving between cells separated by a door.
Given it has no knowledge of the door, nor a skill to open it, the robot's planner will make a plan that will fail during execution (i.e., the robot will be unable to execute its `move' skill between cells separated by a door, as shown in Figure~\ref{fig:overview}(a)).
We assume the robot is equipped with the ability to automatically detect such a failure.
Since the robot has encountered this novelty in its new deployed environment, it can reasonably expect to encounter it in the future when asked to perform similar tasks.
Thus, it must not only efficiently resolve the novelty temporarily to complete the task at hand, but also \textit{learn} such that it can effectively handle such novelty in the future.
Given this, our key question of interest is: how can we efficiently learn to overcome failures due to novelty and solve tasks in our deployment environment?

One straightforward approach to answering our key question is to instantiate a reinforcement learning (RL) problem starting from the state in which the planner fails and provide reward only when the ultimate task goal is achieved.
Unfortunately, such a problem features a continuous action space and sparse reward, which are known to be extremely challenging and sample-inefficient for RL approaches~\cite{lillicrap2015continuous,andrychowicz2017hindsight,bellemare2016unifying}.
Our key idea is that we can make this RL problem significantly easier by exploiting the planner.
In particular, we simply construct the RL problem so that it has access to a special action we call `CallPlanner' that hands control to the planner.
In this way, the policy learned by RL need not to solve the entire long-horizon task, but instead only to resolve the novelty before handing back control to the planner. 
Through the use of the CallPlanner action, our method implicitly learns a policy from the failed state to the $initiation$ $set$ of the planner: the set of states where following the plan returned will bring us closer to the goal state. 
Intuitively, in our running example, the planner already knows how to move and turn on the light: the RL policy must only learn to open the door before handing back control to the planner.
Correctly learning such a ``bridge policy'' allows the agent to smoothly ``bridge'' the gap caused by planning failures. In our example, it also enables the agent to solve tasks involving arbitrary numbers of doors by simply alternating between calling the planner, running the RL policy, and then handing back control to the planner (Figure~\ref{fig:overview} (c)).

We implement our approach in an online learning setting where the agent is provided with a series of tasks in its deployment environment. In experiments, we evaluate the extent to which our approach---\textit{Bridge Policy Learning}---is able to learn policies that can be used in conjunction with planning to overcome novelties encountered during long-horizon tasks.
Experiments across three different simulated environments with varying novelty reveal that our approach is significantly more sample efficient than existing baselines from the literature. We also find that the learned bridge policy, when combined with planning, enables the agent to generalize aggressively to solving more challenging tasks than those encountered during learning.

\section{Problem Setting}
\label{sec:problem-setting}



We consider planning and learning in deterministic, fully-observable environments with object-oriented states. Below, we first describe our assumptions about the environment model before detailing a minimal specification for the planning system, as well as our notion of environment novelty. As a simple running example, we introduce the `Light Switch Door' environment depicted in Figure~\ref{fig:overview}. In this environment, the agent's goal is to turn on the light at the end of a row of grid cells. It possesses skills and a planning model enabling it to move left and right and turn on the light when it is in the corresponding cell. However, at deployment time, it encounters a door that is not part of its planning model. The agent does not possess an explicit skill for opening the door, however, it is able to do this by combining a series of low-level actions.

\subsection{Environment and Task Models}
\label{subsec:environment-models}
Following previous work~\cite{silver2022learningneurosymbolicskillsbilevel}, we take an \textit{environment} to be a tuple $\mathcal{E} = \langle \Lambda, \chi, {U}, f, \Psi \rangle$. $\Lambda$ is the set of possible object types; an object type $\lambda \in \Lambda$ has a name (e.g. \texttt{robot}, \texttt{light}, or \texttt{door} in our running example) and a set of real-valued features (e.g. \texttt{x-position}, \texttt{light-level}) represented as a tuple of continuous values of dimension $\texttt{dim}(\lambda)$.
An object $o$ has a particular name (e.g. \texttt{robby}, \texttt{light-bulb0}) and an associated type (e.g. \texttt{robot}) in any particular problem (i.e., an instance of an environment). An object will specify particular values for each of the features of the associated type (e.g. \texttt{robot} will have \texttt{x-position}$: 0$).
A state $x \in \chi$ is a collection of objects with assigned feature values, and the state space is defined by the set of all objects $\mathcal{O}$, and the possible feature values they could be assigned.
$U$ is the action space consisting of named \textit{parameterized skills}~\cite{da2012learning}, $u(\bar{\lambda}, \circ) \in U$ (e.g. \texttt{MoveRight(robot)}, or \texttt{ToggleLightSwitch(robot, [toggle-value])}, where \texttt{toggle-value} is a continuous parameter).
Each such skill can be executed by specifying the list of objects with types $\bar{\lambda}$, which causes the environment to advance to a new state $x' \in \chi$ according to some unknown transition model $f: \chi \times U \rightarrow \chi$.
We assume there always exists a particular skill named \texttt{RunLowLevelAction}, which takes in no objects and whose continuous parameters provide access to a very low-level action space (e.g. the robot's motor commands) compared to other skills.
Finally, $\Psi$ is a set of \textit{predicates}.
Each \textit{predicate} $\psi \in \Psi$ has a name (e.g., \texttt{Light-On}) and a tuple of types (e.g., (\texttt{light})).
\textit{Grounding} a predicate yields a \emph{ground atom}, which is a predicate and a mapping from its type tuple to objects (e.g., \texttt{Light-On}(\texttt{light-bulb0})).
Given a particular state $x \in \chi$, a ground atom can be run on this state to produce a boolean value.
Predicates induce a state abstraction: $\textsc{abstract}(x)$ denotes the set of ground atoms that hold true in $x$, with all others assumed false.
We denote a set of ground atoms via $\bar{\psi}$.
We denote the abstract state as $s$ (i.e., $s = \textsc{abstract}(x)$).

Each environment is associated with some \textit{task distribution} $T$.
A \textit{task} $t \in T$, is a tuple $\langle \mathcal{O}, x_0, G, H \rangle$, where $\mathcal{O}$ is some set of objects and $x_0 \in \chi$ is an initial state.
$G$ is a collection of ground atoms describing the goal (e.g. [\texttt{Light-On}(\texttt{light-bulb0})]), and $H$ is a maximum horizon (i.e., number of actions) within which the task must be solved.
A solution to a task is sequence of actions $u_0, u_1, \ldots, u_n$, such that $n < H$ and taking these actions from the initial state yields a final state $x_n$ such that the goal expression holds (i.e. $\forall g \in G, g(x_n)$).

\subsection{Planning and Environment Novelty}
\label{subsec:planning}
We assume the agent has access to a planner $P$ which takes a task $t$ and corresponding environment $\mathcal{E}$ as input and produces a policy $\pi_{\text{plan}}$.
Importantly, this planner operates over some internal model that \textit{does not} account for environment novelty.
This policy in turn takes in a state, and outputs (1) an action to take, (2) a boolean `stuck' indicator (i.e., $\pi_{\text{plan}}: \chi \rightarrow U \times \{0, 1\}$).
The stuck indicator can be interpreted as a simple measure of the robot's confidence in its output: when it is false, the robot believes its action will make progress towards the goal as intended, whereas when it is true, the policy believes some novelty has been encountered that it is not equipped to handle.
For example, in a problem from the `Light Switch Door` in which there are 3 cells, a door between the second and third cells, and a light switch in the final cell, the planner will produce a policy that outputs a `MoveRight' action with the stuck indicator set to false in the first and second cells, and then produce any action with the stuck indicator set to true after failing to move through the door~\footnote{Internally, the policy has a model of the expected effects of each action (computed by the planner), and realizes that it is stuck after these effects do not hold}.
For the purposes of our problem, we can treat this planner and its output policy largely as black boxes, and thus we defer discussing implementation details to Appendix~\ref{app:planner-details}.
We refer to the state from which the stuck indicator is first set to true as the stuck state $x_{\text{stuck}}$, and are primarily interested in learning  to complete the corresponding task from it.

We consider an online learning setting in which to learn to overcome this novelty. We measure both the task solves achieved during learning process and the performance on evaluation tasks after each learning update.
Specifically, \textit{tasks} are drawn from $T$.
The agent has a finite budget of environment steps $H_{\text{train}}$ every \textit{episode} (i.e., before the environment is reset).
Our objective is to solve as many tasks drawn from this distribution $T$ as possible given a limited time horizon for each task.
For the purposes of measuring generalization, in Section~\ref{sec:experiments} we artificially split tasks into \textit{training} and held-out \textit{evaluation} tasks such that the evaluation tasks are constructed to have similar goals, but often involve more objects (e.g. more cells with more doors in the case of the Light Switch Door environment), requiring the agent to generalize beyond the training distribution.

\section{Bridge Policy Learning}
\label{sec:method}



How should the agent formalize and optimize the objective introduced in Section~\ref{subsec:planning}?
A simple but naive approach would be to simply behave randomly for a random amount of time and then try to follow the plan again.  
However, this process would need to be repeated for every new problem, and would likely lead to the task horizon being exhausted in most cases.
Another approach would be to simply perform RL in a new Markov Decision Process (MDP) that models the environment, task, and novelty.
However, not only would this approach need to learn to overcome whatever novelty got the planner stuck, it would also need to complete the rest of the task.
This is likely to be quite sample inefficient.
In particular, it does not leverage or exploit knowledge the planner already possesses.
For instance, in our running example from the Light Switch Door environment, the planner would be able to achieve the task goal once the door is opened: we only need to learn how to open the door.

We propose to learn a \textit{bridge policy} that the agent can use during execution to get to a state where replanning and following the new plan will no longer get stuck.
At evaluation time, the agent will start by planning, and then executing its plan until it gets stuck.
It will then use its bridge policy until this policy terminates when it calls the planner, then replan and execute the new plan once again. 
This iteration between the planner and the bridge policy continues until the agent achieves the task goal or the task horizon is exhausted.
To learn such a bridge policy, we need to define a set of target states for the policy to reach.
In this work, we propose to have the agent automatically discover such a set of states implicitly.
We do this by setting up an RL problem in which one of the available actions is to hand back control of the agent to the planner, so that the agent learns the optimal state to start replanning.

Below, we first describe the construction of the RL problem, then discuss how we choose to solve it. Finally, we describe how we use the learned policy at evaluation time to generalize to problems in our evaluation set.

\subsection{Constructing an RL Problem}
\label{subsec:construct-mdp}
Once the agent detects that it is in a `stuck' state $x_{\text{stuck}}$, we set up an RL problem by constructing an MDP $\mathcal{M} = \langle \chi', U', f, R, \gamma \rangle$.
Here, $\chi'$ represents the RL problem's state space, $U'$ its action space, $f$ its transition function (which is the same as the original task's environment), $R$ its reward function, and $\gamma$ its discount factor.

The state space $\chi'$ is a subspace of the corresponding environment's state space $\chi$. In particular, we construct $\chi'$ by selecting a subset of the objects $\mathcal{O}$ that make up $\chi$.
The purposes of this feature selection are twofold: to improve the sample efficiency of RL, and to improve the generalization of the learned bridge policy.
Many possible strategies for selecting such a subset are possible (e.g. \cite{silver2021planning}), but in this work, we adopt the simple strategy of selecting the single object that can be interacted with that is closest to the robot object by euclidean distance in $x_{\text{stuck}}$.
In our running example, this will simply be the `door' object.
Learning a bridge policy that's specific to this object's state enables this policy to be reused in novel problems that may include additional variations in different objects (e.g. problems with more doors or grid cells in our running example).

The action space $U'$ is simply the original environment's action space augmented with an extra action we call `\textit{CallPlanner}' (i.e. $U' = U \cup \{\text{CallPlanner}\}$)~\footnote{In environments where it is possible for the robot to take unsafe actions, it would be useful to further restrict or modify this action space, but that is beyond the scope of this current work.}.
As the name suggests, this action simply returns control of the agent to the planner: the agent begins executing a sequence of actions output by the planner, and terminates in a new state $x'$ where either (1) the goal is achieved, (2) the task horizon has expired, or (3) the planner is stuck again.
Importantly, even though the the planner itself executes many actions before terminating, we treat the entire CallPlanner sequence as one atomic action within the MDP $\mathcal{M}$.

Finally, the reward function is simply a sparse reward corresponding to achieving the task goal $G$.
$$R(x_t, a_t)=\begin{cases} 1 &\text{   if $G \subseteq$ \textsc{abstract}($x_{t+1}$)}, \\ 0 &\text{ otherwise }\end{cases}.$$
We set the discount factor $\gamma$ to be less than $1$ so that the agent is encouraged to solve the MDP with the fewest possible actions. Once the robot gets to a state it can plan from, executing the CallPlanner atomic action will bring the robot to the goal in one step, so an optimal bridge policy would take the fewest possible actions to resolve the novelty before calling the planner.

As is standard for RL problems, the learning objective is to learn a policy $\pi_{\text{bridge}}$ to maximize the discounted sum of rewards in expectation over all MDPs that comprise our training distribution:
$$\mathbb{E}_{\mathcal{M} \sim T} \sum_{\substack{t=0 \\ a_t \sim \pi(x_t)}}^{H} \gamma^{t} R(x_t, a_t).$$

\subsection{Learning a Bridge Policy}
\label{subsec:solving-mdp}

\begin{wrapfigure}{L}{0.6\textwidth}
    \begin{minipage}{0.6\textwidth}
        \begin{algorithm}[H]
        \caption{Approach}
        $\textbf{input: }\text{Environment } \mathcal{E} = \langle \Lambda, \chi, {U}, f, \Psi \rangle, \text{Task } t = \langle \mathcal{O}, x_0, G, H \rangle$, Goal $G$, Planner $P$, Bridge Policy $\pi_{\text{bridge}}$.
        \begin{algorithmic}[1]
        \State $\pi_{\text{plan}} = P(t, \mathcal{E})$ \Comment{Start by planning}
        \State $x = x_0$
        \State num\_steps $= 0$
        \While{$G \subsetneq \textsc{abstract}(x)$ and num\_steps $< H$}
            \State $u, \text{stuck} = \pi_{\text{plan}}(\textsc{abstract}(x))$
            \Comment{Follow plan}
            \While{stuck $!= 1$}
                \State $x = f(u, x)$
                \State $u, \text{stuck} = \pi_{\text{plan}}(\textsc{abstract}(x))$
                \State num\_steps $+= 1$
            \EndWhile
            \While{u $!=$ `CallPlanner'}
                \State $u = \pi_{\text{bridge}}(x)$ \Comment{Follow bridge until CallPlanner}
                \State $x = f(u, x)$
                \State num\_steps $+= 1$
            \EndWhile
        \EndWhile
        \end{algorithmic}
        \label{alg:rlbridgepolicysolve}
        \caption{Our overall meta-policy for solving tasks by leveraging a planner $P$ and a learned bridge policy $\pi_{\text{bridge}}$}.
        \end{algorithm}
    \end{minipage}
\end{wrapfigure}

Given the MDPs constructed in Section~\ref{subsec:construct-mdp}, we can apply any RL algorithm capable of handling continuous states and actions to learn a policy $\pi_{\text{bridge}}$.
However, in our setting, actions take the form of parameterized skills~\footnote{Thus, our MDP is actually a parameterized action MDP (PAMDP)~\cite{pamdp}}.
We thus take advantage of this by leveraging recently-introduced efficient architectures for RL with parameterized skills~\cite{nasiriany2022maple,kumar2024practice}.
Specifically, we choose to use a variant of the pure RL approach introduced in~\cite{kumar2024practice}.

Following this approach, which is related to Deep Q Learning \cite{mnih2013playingatarideepreinforcement}, we learn a Q-function $Q(x, a)$ that maps MDP states and skills (with all parameters specified) to Q-values.
At evaluation time, we sample $n_{\text{sample}}$ fixed number of continuous parameters uniformly at random for each skill among our set of actions $U'$ and pick the argmax.
To prevent overestimation of Q-values, we maintain a separate target network in accordance with Double DQN~\cite{vanhasselt2015deepreinforcementlearningdouble}.

We randomly initialize our Q-networks, and collect data for training them by performing epsilon-greedy exploration with annealing of the epsilon parameter. Here, epsilon greedy intuitively serves to balance how much we value solving the current task with how much we want to explore and learn new things about our environment.






\subsection{Using a Learned Bridge Policy}
\label{subsec:bridge-eval}
After training our bridge policy $\pi_{\text{bridge}}$ we can leverage it to recover from novelties encountered during evaluation time by switching control of the agent between using its planner and using its learned bridge policy.
Our meta policy for solving tasks via both planning and using the learned bridge policy is shown in Algorithm~\ref{alg:rlbridgepolicysolve}.
Given a task, we first call the planner and execute its plan until a `stuck' state is reached (if one is reached at all).
We then pass control to the bridge policy, which will perform some sequence of actions before either achieving the task goal, or invoking its CallPlanner action to hand control of the agent back to the planner.
In this way, a bridge policy that has correctly learned to overcome an encountered novelty can be used by the agent to generalize to testing tasks involving more instances of that novelty (assuming every instance can be resolved by the same policy) by calling the same bridge policy multiple times in sequence whenever the novelty is encountered.
We evaluate this capability empirically in Section~\ref{sec:experiments} below.

\section{Experiments}
\label{sec:experiments}

\begin{figure*}[t]
\centering
\subfigure[Light Switch Door]{
\scalebox{0.25}{\includegraphics[]{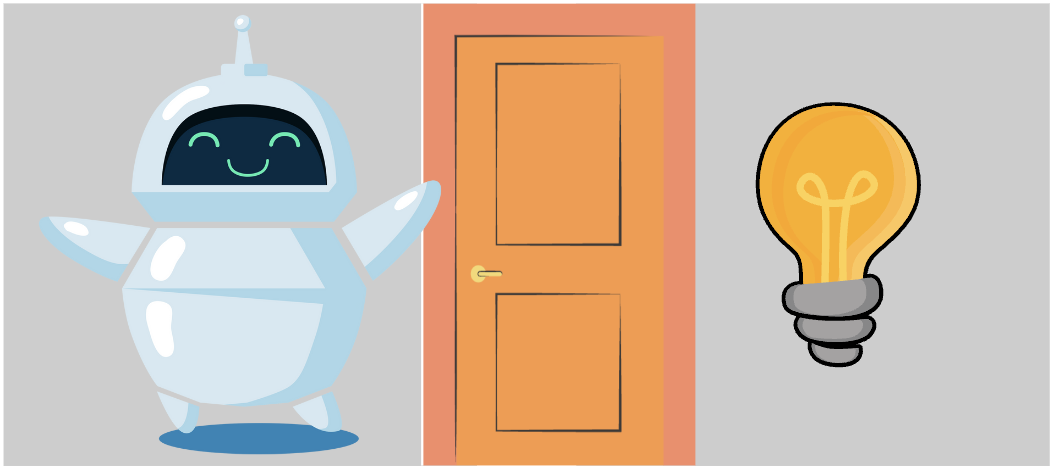}}}
\centering
\subfigure[Doorknobs]{
\scalebox{0.11}{\includegraphics[]{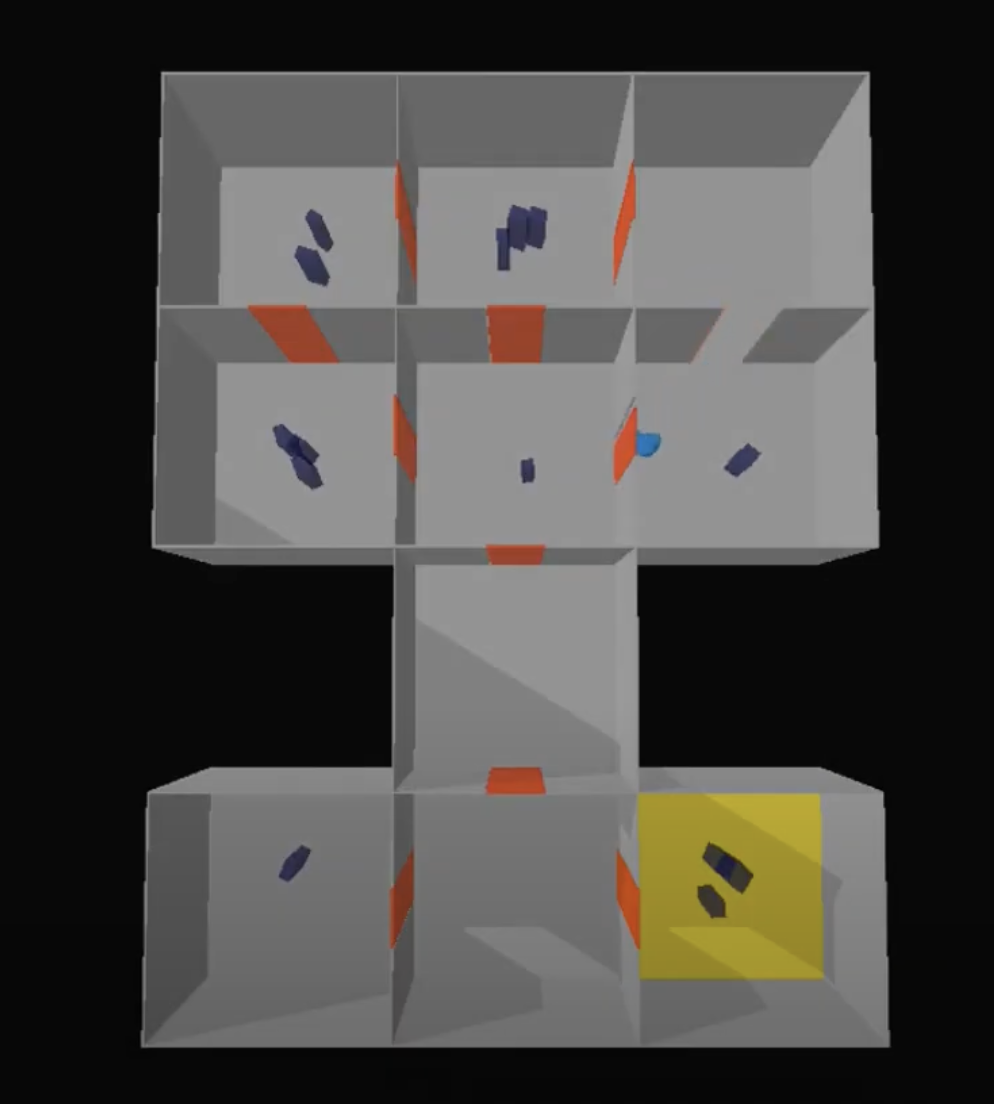}}
}
\centering
\subfigure[Coffee]{
\scalebox{0.3}{\includegraphics[]{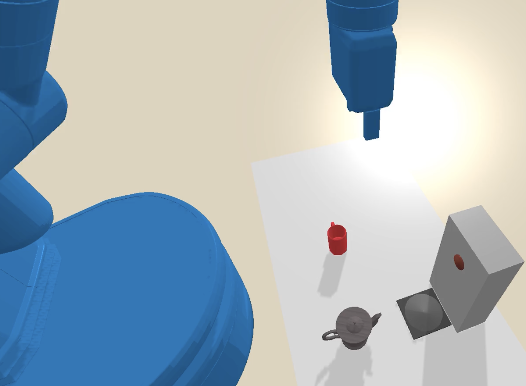}}
}
\label{envs}
\caption{\small{Visualizations of our different experimental domains.}}
\vspace{-1.5em}
\end{figure*}

Our experiments are designed to answer the following questions about our approach:
\begin{tightlist}
    \item[\textbf{Q1.}] How sample efficient is bridge policy learning compared to alternatives?
    \item[\textbf{Q2.}] How well can our learned bridge policy be used to generalize to harder versions of the tasks?
    \item[\textbf{Q3.}] How important is the $CallPlanner$ action in our approach?
\end{tightlist}

\textbf{Environments: } We provide high-level descriptions of our three simulated environments here.

\begin{tightlist}
\item \textit{Light Switch Door}: An implementation of our toy running example depicted in Figure~\ref{fig:overview}. The robot must traverse a row of cells to turn on a light at the end of the row. However, there are doors obstructing the way that the robot's planner does not model, and thus following its output plan will result in a stuck state. The robot needs to learn how to open these doors using two low level actions.
This domain is a variation on `Light Switch' from~\cite{kumar2024practice}. 


\item \textit{Doorknobs}: Equivalent to the `Doors' environment from~\cite{silver2022learningneurosymbolicskillsbilevel}. Here, a robot must navigate to a target room while avoiding obstacles and opening doors. The planner has no direct knowledge of the existence of doors, nor how to open them. A skill that moves the robot between configurations using a motion planner (BiRRT) is provided; door opening must be learned by a bridge policy. Tasks have 4--25 rooms with random connectivity. The goals, obstacles, and initial robot pose also vary.


\item \textit{Coffee}: Another environment taken directly from~\cite{silver2022learningneurosymbolicskillsbilevel}. The robot makes coffee by putting the jug in the coffee machine, turning the machine on, and then pouring the coffee into a cup. However, the jug is initially rotated in a way so that the robot must first rotate the jug in order to grasp it. Otherwise, if the robot tries to grasp the jug without first rotating it, it will enter a stuck state.

\end{tightlist}

\textbf{Approaches: } We now briefly describe the various approaches we evaluate in the above environments. 
\begin{tightlist}
    \item \textit{Bridge Policy Learning: } Our main approach.
    \item \textit{Ours no feature selection: } An ablation of our approach that does not perform any feature selection when constructing the RL problem, but instead passes in the entire task state (i.e., $\chi' = \chi$). Note that we do not test this ablation in the Doorknobs environment because the entire state is too large and the method times out after several learning cycles.
    \item \textit{Ours no CallPlanner: } An ablation of our approach that does not include CallPlanner in the RL problem's action space. This ablation is not able to exploit the planner's knowledge.
    \item \textit{Random Bridge: } Do not learn using the training tasks; behave randomly (including attempting to call the planner) whenever the planner reaches a stuck state.
    \item \textit{Pure Planning: } Do not learn using the training tasks; keep attempting to call the planner whenever the planner reaches a stuck state.
    \item \textit{Maple Q: } A pure RL approach from~\cite{kumar2024practice} inspired by~\cite{nasiriany2022maple}. Do not use the planner at all, not even at the start. Rather, attempt to learn a policy to achieve the goal from the initial state for each of the training tasks. There is also no state space feature selection in this method.
\end{tightlist}

\begin{figure*}[t]
\centering
\subfigure{
\scalebox{0.28}{\includegraphics[]{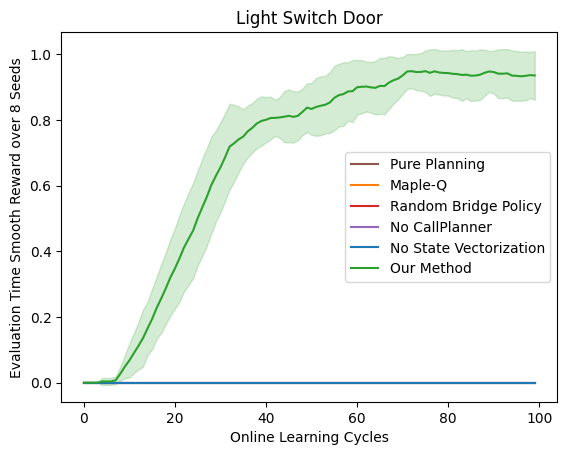}}
}
\subfigure{
\scalebox{0.28}{\includegraphics[]{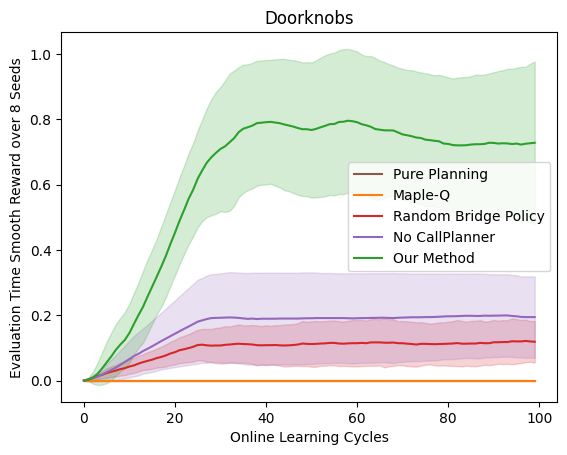}}
}
\centering
\subfigure{
\scalebox{0.28}{\includegraphics[]{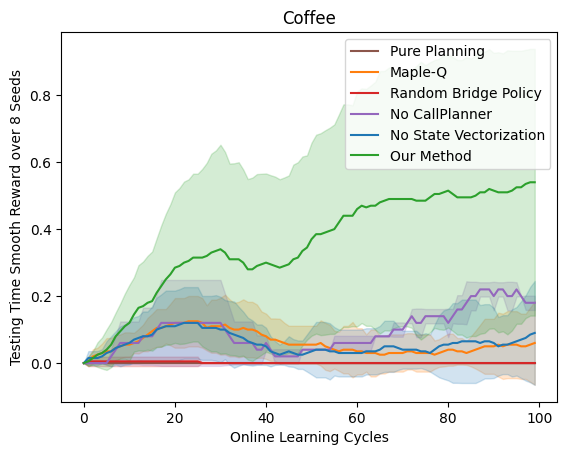}}
}
\subfigure{
\scalebox{0.28}{\includegraphics[]{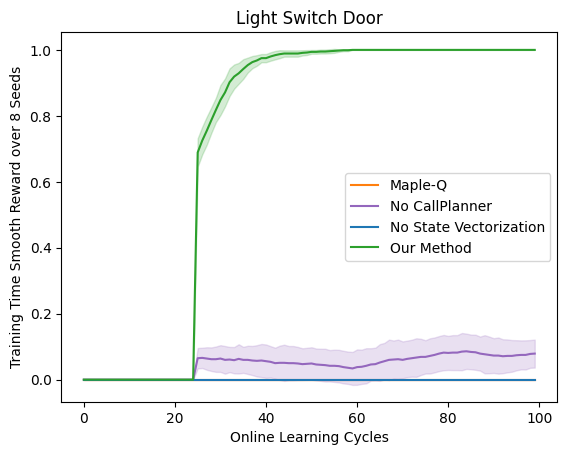}}
}
\centering
 \subfigure{
\scalebox{0.28}{\includegraphics[]{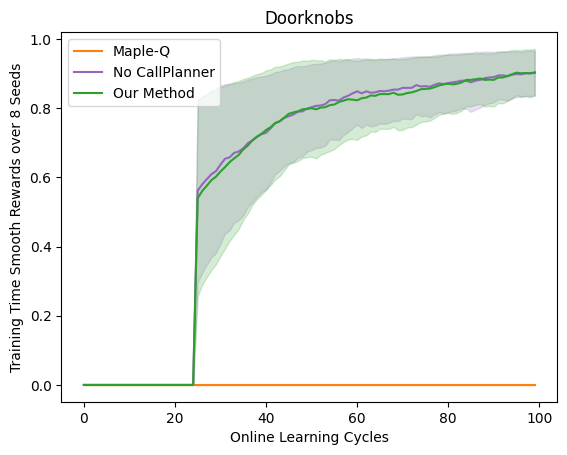}}
}
\subfigure{
\scalebox{0.28}{\includegraphics[]{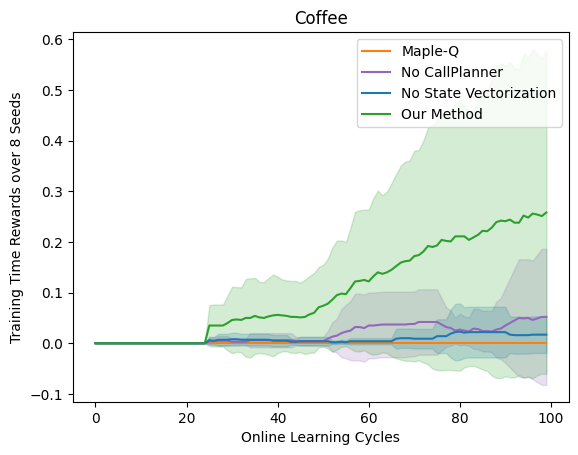}}
}

\label{fig:exp}
\vspace{-1em}
\caption{\small{Experiments in our different domains. Results from both evaluation time and training time. We average smooth reward, the reward averaged over the last 25 online learning cycles, across all seeds and graph the variance.}}
\vspace{-1em}
\end{figure*}

\textbf{Experimental Setup: } We run 8 independent seeds of each approach in each environment. Different seeds yield different training and test problems where the numbers of objects and initial states of the task are randomized. We fix the train and evaluation horizons to be the same for all approaches in each environment. Our key quantitative metrics of interest are smooth test time reward, and smooth training time reward, to track the learning process of the agent over time. Additional details on hyperparameters and experiments are presented in Appendix~\ref{app:additional-method-details}.

\textbf{Results and Analysis:} Figure ~\ref{fig:exp} shows that our method strongly outperforms the baselines across environments. Specifically, our method is much more sample efficient, converging at around the $30$th online learning cycle when it has learned from and collected only $150$ trajectories in Light Switch Door and Doorknobs. Our approach achieves significantly higher training time reward in Coffee and Light Switch Door than all other methods. In Doorknobs, the ablation without the CallPlanner action is also able to reach high training rewards because the environment in those experiments is very simple with only 2x2 grids. In this environment, the ablation can simply learn the full trajectory to the goal without calling the planner. When the ablation is evaluated in a larger environment, however, our method obtains much higher reward in comparison. Without CallPlanner, we find that the ablation simply gets ``stuck'' after overcoming the first novelty during evaluation time in all three environments.  Our results show that the ablation without feature selection is not able to learn as efficiently. CallPlanner and feature selection are both critical aspects of our approach. 
\section{Related Work}
\label{sec:related-work}
\textbf{Open-World AI: } There has been a long history of work that attempts to develop agents capable of continuously adapting to novel scenarios~\cite{oudeyer2007intrinsic,pathak2017curiosity, colas2019curious, santucci2020intrinsically, 10160382, GOEL2024104111}.
Many of these works~\cite{oudeyer2007intrinsic,pathak2017curiosity, colas2019curious, santucci2020intrinsically} are interested in a setting where the agent is not provided any models nor specific goals a priori but rather must simply explore in an open-ended fashion.
By contrast, our work is interested in more directed exploration, where the agent is specifically trying to overcome some failure and also is equipped with a model and planner to be leveraged.
The works that operate in a setting similar to ours~\cite{10160382, GOEL2024104111} typically focus on two aspects: novelty detection, and novelty adaptation.
Additionally, these works often operate in environments with purely discrete action spaces. 
Our work assumes a relatively simple form of novelty detection and focuses on a particular form of novelty caused by a particular skill failing to achieve its affect due to particular object(s) in the environment.
However, we operate in environments with fundamentally continuous action spaces, and make no assumptions about any existing recovery strategies.

\textbf{Learning new skills for planning:} Previous works have studied adding or adapting skills for a planning model by learning from expert demonstrations~\cite{10252153, silver2022learningneurosymbolicskillsbilevel}, and from online interaction~\cite{mendez-lifelong,league,goel2022rapidlearnframeworklearningrecover,10160382}.
Similar to our approach, works that learn from online interaction generally assume the agent is following a provided plan until an error is encountered, at which point it switches to RL. Importantly, they assume the original computed plan is correct (but for the error) and thus aim to learn to resolve the error and simply continue the plan. This assumption enables them to leverage the original plan for reward shaping. By contrast, our approach makes no explicit commitment to the original plan, but rather enables the agent to replan after the learned policy has terminated. This enables our approach to tackle a wider range of environments as demonstrated experimentally in Section~\ref{sec:experiments} by the ablation that does not contain the CallPlanner action~\footnote{We do not directly compare to baselines that leverage the original plan for reward shaping (e.g. \cite{goel2022rapidlearnframeworklearningrecover, 10160382})}. Specifically, the set of states reachable from the planner that these methods learn is a subset of the states we learn. Therefore, our method is more general and able to find a more optimal trajectory in certain tasks. However, we must rely on only the sparse reward of achieving the agent's goal, and thus could be significantly less sample-efficient in complex environments. Intuitively, previous methods using RL can be thought of as giving a planner ``access'' to learning an RL policy. Our method builds upon this by also doing the converse: giving the RL agent access to a planner.

\textbf{Recovering from failures in robotics: } Previous approaches in diagnosis and plan repair for robotics have handled recovery via pre-specified heuristics~\cite{fault-detection, behavior-trees-faults, 813040} or prompting a human user~\cite{innes2020elaboratinglearneddemonstrationstemporal, tellex2014asking}.
More recently, several methods~\cite{goel2022rapidlearnframeworklearningrecover, kumar2024practice} seek to autonomously \textit{adapt} existing skills to succeed under unexpected environmental changes. 
This assumes that there exists a skill in the robot's repertoire that can be adapted to cope with the encountered novelty.
We do not make this assumption, but rather learn a novel recovery strategy that may be composed of existing skills and low-level actions.
\section{Conclusion}
\label{sec:conclusion}
In this work, we proposed a method that enables a robot equipped with a model-based planner to adapt to novel deployment environments where following the returned plan fails to complete an assigned task.
Experiments in simulated domains revealed that our method is able to successfully overcome this novelty with very low sample complexity and generalize to larger environments.

There are several limitations of our approach.
Firstly, we rely on reducing the dimensionality of the input state space to enable generalization of the bridge policy to novel, more complex tasks. We currently use a very simple approach for this that is unlikely to work in more complex and realistic environments.
Secondly, we attempt to learn a single bridge policy that is capable of coping with all encountered novelties.
This suffices for environments with very few novelties that lead to planning failure, but would not work in environments featuring multiple different types of novelty.
Finally, our approach performs a relatively simple form of exploration (epsilon-greedy), which works well when the desired recovery behavior is relatively short horizon, but will likely not scale well to settings that require learning a much more long-horizon bridge policy.

In the future, we hope to address these limitations and enable our approach to be realized in complex and useful robotics domains. An important direction is to integrate our approach with perception to perform decision making from camera input, both for our planner~\cite{kumar2024practice}, and for our learned bridge policy~\cite{nasiriany2022maple}.
In conjunction with this, integrating pretrained vision language models (VLMs) could help perform automatic dimensionality reduction for our bridge policy by suggesting which objects to ignore or consider after a stuck state is encountered.
This could also be directly learned via a graph neural network (GNN), as demonstrated in previous work~\cite{silver2021planning}.
Additionally, it will be important to run our approach in a wider range of complex environments, and compare directly to closely-related approaches from the literature (e.g. \cite{goel2022rapidlearnframeworklearningrecover,10160382}). This would allow us to further test the efficacy of our CallPlanner formulation in comparison to other approaches that have achieved high performance in similar problem settings.

\bibliographystyle{IEEEtran}
\bibliography{IEEEabrv,references}

\clearpage
\appendix
\section{Additional Experimental Details}

\label{app:additional-method-details}


\begin{table}[!h]
\centering
\caption{Environment Hyperparameters}
\label{tab:env_hyperparameters}
\vspace{0.5em}
\begin{tabular}{lccc}
\toprule
Env & Light Switch Door & Doorknobs & Coffee \\
\midrule
Horizon & max(30, h+5) & 200 & 100 \\
Steps per learning trajectory & 100 & 100 & 100 \\
Trajectories per learning cycle & 5 & 5 & 5 \\
Number of training tasks & 1 & 1 & 1 \\
Number of test tasks & 10 & 10 & 1 \\
Number of test novelties & 2--4 & 2--5 & 1 \\
Number of test cells & 10--20 & 4--25 & NA \\
\bottomrule
\end{tabular}
\vspace{0.5em}
\begin{flushleft}
\small{Note: $h$ is the minimum horizon length needed to complete the task in Light Switch Door.}
\end{flushleft}
\end{table}

\begin{table}[!h]
\centering
\caption{Method Hyperparameters}
\label{tab:research_parameters}
\vspace{0.5em}
\begin{tabular}{ll}
\toprule
Parameter & Value \\
\midrule
Discount factor $\gamma$ & 0.8 \\
Learning rate & $10^{-3}$ \\
Polyak averaging & $2.5^{-3}$ \\
$\epsilon$ annealing reduction per step & $3.8 \cdot 10^{-5}$ \\
MLP hidden layer sizes & [32, 32] \\
MLP max iters & $10^4$ \\
Replay buffer size & $10^6$ \\
Optimizer & Adam \\
Weight decay & $0$ \\
\bottomrule
\end{tabular}
\vspace{0.5em}
\end{table}

All experiments were run on Intel Xeon Gold 6130 Processor CPUs.

\section{Planner Implementation Details}
\label{app:planner-details}
Here, we provide additional details about the implementation of the planner used in this work.
We note that this is a relatively simple implementation that satisfies the setup in Section~\ref{sec:problem-setting} and other, more sophisticated implementations are possible as well.

We adopt a planning implementation that is identical to that of~\cite{kumar2024practice}.
This requires access to a set of extra predicates, $\Psi_{\text{planner}}$ in addition to those specified as part of the environment.
In everything that follows, assume that when we perform abstraction (i.e., calling \textsc{abstract}), we will use the environment predicates $\Psi$ as well as the planner predicates $\Psi_{\text{planner}}$.
For example, in our toy Light Switch Door environment, the environment predicates will include only `LightOn(light)', while the agent will additionally have predicates such as `RobotInCell(robot, cell)' and `Adjacent(?c1:cell, ?c2:cell)'.
Next, we assume access to PDDL symbolic planning operators~\cite{fox2003pddl2} with predicate-based preconditions and effects~\footnote{Previous work~\cite{silver2021learning,silver2023predicate,kumar2023learning} has learned operators and predicates; but we choose to manually specify them here}.
For example, the operator \texttt{Move(robot},~\texttt{current\_cell},~\texttt{target\_cell}~\texttt{)} is:
\clearpage
\begin{Verbatim}[frame=single,resetmargins=true]
    MoveRobot(robot, current_cell, 
                target_cell)
    :precondition (and 
      (Adjacent current_cell 
                    target_cell)
      (RobotInCell robot current_cell)) 
    :effect (and
      (RobotInCell robot current_cell)
      (not (RobotInCell robot 
            target_cell)))
\end{Verbatim}

Importantly, note that these symbolic operators are entirely discrete and do not feature continuous parameters.
Here, the preconditions characterize the initiation conditions of the skill.
For example, in the above operator, we say the preconditions hold in a state $x$ if $\{\texttt{Adjacent(current\_cell, target\_cell)},$ $\texttt{RobotInCell(robot, current\_cell)}\} \subseteq \textsc{abstract}(x)$.
Similarly, the effects characterize aspects of the state we expect to be true after the skill is executed.
We say the operator has been executed successfully in a state $x'$ (where $x'$ is the result of executing a skill $u$ from the state $x$; $x' = f(x, u)$) if $\{\texttt{RobotInCell(robot, current\_cell)}\} \subseteq \textsc{abstract}(x')$ and also if $\{\texttt{RobotInCell(robot, target\_cell)}\}$ $\not\subset \textsc{abstract}(x')$.

We assume that each operator is linked with a (1) parameterized skill from our environment's action space $U$, as well as (2) a \textit{parameter policy} that takes in an environment state $x$ and outputs a distribution over continuous parameters for the associated skill.

Given a task (Section ~\ref{sec:problem-setting}), we use the initial state $x_0$ and a goal $G$, we construct a PDDL planning problem with initial state $\textsc{abstract}(x_0)$ and use a PDDL planner~\cite{helmert2006fast} to efficiently generate a sequence of planning operators (a skeleton) that chain together by precondition and effect to reach the goal.
In experiments, we perform $A^*$ search using the LM-Cut heuristic.

Once a skeleton is obtained, we construct the planner policy $\pi_{\text{plan}}$ such that it sequentially executes the skill associated with each operator in the plan by greedily selecting parameters from the operator's associated parameter policy. We implement the boolean stuck indicator by simply checking whether the subsequent operator's preconditions hold in the state resulting after executing this skill with the provided continuous parameters.

\end{document}